\documentclass{ieeetj}
\usepackage{cite}
\usepackage{amsmath,amssymb,amsfonts}
\usepackage{algorithmic}
\usepackage{graphicx,color}
\usepackage{textcomp}
\usepackage{xcolor}
\usepackage{hyperref}
\hypersetup{hidelinks=true}
\usepackage{algorithm,algorithmic}
\def\BibTeX{{\rm B\kern-.05em{\sc i\kern-.025em b}\kern-.08em
    T\kern-.1667em\lower.7ex\hbox{E}\kern-.125emX}}
\AtBeginDocument{\definecolor{tmlcncolor}{cmyk}{0.93,0.59,0.15,0.02}\definecolor{NavyBlue}{RGB}{0,86,125}}

\def\OJlogo{}
\def\seclogo{}

\usepackage{relsize}
\usepackage{xspace}
\usepackage{color}
\usepackage{graphics}
\usepackage{graphicx}

\usepackage{amsmath}

\usepackage{amsfonts}
\usepackage{amssymb}
\usepackage{amsthm}

\usepackage{float}
\usepackage{url}
\usepackage{booktabs}
\usepackage{multirow}
\usepackage{hhline}
\usepackage[capitalise]{cleveref}
\usepackage{dsfont}
\usepackage{cancel}

\usepackage{subcaption}
\usepackage{catchfilebetweentags}

\usepackage{booktabs}

\usepackage{csvsimple}

\usepackage{tikz}
\usetikzlibrary{positioning}

\usepackage{pgfplots}
\pgfplotsset{compat=1.18}  

\usepackage{bm}

\newcommand{\bdmath}{\begin{dmath}}
\newcommand{\edmath}{\end{dmath}}
\newcommand{\beq}{\begin{equation}}
\newcommand{\eeq}{\end{equation}}
\newcommand{\bdm}{\begin{displaymath}}
\newcommand{\edm}{\end{displaymath}}
\newcommand{\bea}{\begin{eqnarray}}
\newcommand{\eea}{\end{eqnarray}}
\newcommand{\beal}{\beq \begin{array}{ll}}
\newcommand{\eeal}{\end{array} \eeq}
\newcommand{\beas}{\begin{eqnarray*}}
\newcommand{\eeas}{\end{eqnarray*}}
\newcommand{\ba}{\begin{array}}
\newcommand{\ea}{\end{array}}
\newcommand{\bit}{\begin{itemize}}
\newcommand{\eit}{\end{itemize}}
\newcommand{\ben}{\begin{enumerate}}
\newcommand{\een}{\end{enumerate}}

\newcommand{\hide}[1]{}

\newcommand{\hiddenText}{{\color{gray} hidden text.}}
\newcommand{\hideWithText}[1]{\hiddenText}

\newcommand{\blue}[1]{{\color{blue}#1}}

\newcommand{\linkToPdf}[1]{\href{#1}{\blue{(pdf)}}}
\newcommand{\linkToPpt}[1]{\href{#1}{\blue{(ppt)}}}
\newcommand{\linkToCode}[1]{\href{#1}{\blue{(code)}}}
\newcommand{\linkToWeb}[1]{\href{#1}{\blue{(web)}}}
\newcommand{\linkToVideo}[1]{\href{#1}{\blue{(video)}}}
\newcommand{\award}[1]{\xspace} 

\begin{document}
\receiveddate{}
\reviseddate{}
\accepteddate{}
\publisheddate{}
\currentdate{}

\markboth{}{Lajoie {et al.}}

\title{Multi-Robot Decentralized Collaborative SLAM in Planetary Analogue Environments: Dataset, Challenges, and Lessons Learned}

\author{Pierre-Yves Lajoie, Karthik Soma, Haechan Mark Bong, Alice Lemieux-Bourque, Rongge Zhang, Vivek Shankar Varadharajan, Giovanni Beltrame}
\affil{The authors are with the Department of Computer and Software Engineering, Polytechnique Montreal, Montreal, Canada}
\corresp{Corresponding author: Pierre-Yves Lajoie (email: pierre-yves.lajoie@polymtl.ca).}
\authornote{This work was partially supported by a Vanier Canada Graduate Scholarships Award, the Fonds de recherche du Québec - Nature et technologies, and by the Canadian Space Agency.}

\begin{abstract}
    Decentralized Collaborative Simultaneous Localization and Mapping (C-SLAM) is
essential to enable multi-robot missions in unknown environments without
relying on pre-existing localization and communication infrastructure. This
technology is anticipated to play a key role in the exploration of the Moon,
Mars, and other planets. In this paper, we share insights and lessons
learned from C-SLAM experiments involving three robots operating on a Mars
analogue terrain and communicating over an ad-hoc network. We examine the impact
of limited and intermittent communication on C-SLAM performance, as well as the
unique localization challenges posed by planetary-like environments.
Additionally, we introduce a novel dataset collected during our experiments,
which includes real-time peer-to-peer inter-robot throughput and latency
measurements. This dataset aims to support future research on
communication-constrained, decentralized multi-robot operations.

\end{abstract}


\maketitle

\section{INTRODUCTION}
\label{sec:introduction}

\ExecuteMetaData[explanation_figures.tex]{csa-field}

\IEEEPARstart{M}{ulti}-robot systems hold the potential to revolutionize
space and planetary exploration. Teams of robots can parallelize work, be more
resilient to individual failures, and, most importantly, collaborate to
accomplish collective tasks that are out of reach of single-robot systems.
However, operating robots on another planet presents some unique challenges,
such as large communication delays with base stations on Earth, difficult and
unknown terrain, or the absence of any pre-existing infrastructure. In these
conditions, robots need high levels of autonomy to operate safely.

One of the key enablers of robot autonomy is the Simultaneous Localization and
Mapping (SLAM) algorithm~\cite{cadenaPresentFutureSimultaneous2016}, which
provides localization estimates of the robot and a map of the surrounding
environment that can be used for terrain analysis, path planning, and decision
making. In the case of multi-robot systems, the robots need to collaborate in
the localization and mapping process in order to converge to a consistent
perception of the environment across the team of robots. Without shared
situational awareness, individual robots are constrained by their limited view
of the environment and are not able to collaborate efficiently. Thus,
Collaborative SLAM
(C-SLAM)~\cite{saeediMultipleRobotSimultaneousLocalization2016,lajoieCollaborativeSimultaneousLocalization2022}
is likely to be a vital component of future multi-robot missions on the Moon,
Mars, or other planets.

That being said, there are additional key requirements for the efficient
deployment of C-SLAM algorithms in space. First and foremost, due to the high
cost of space missions, the systems need to be as robust and resilient to
failures as possible. Also, due to the lack of pre-existing localization and
networking infrastructure, individual robots need enough autonomy from any base
station (on Earth or on the explored planet itself) to survive frequent and
lengthy disconnections over the course of their missions. Given those
constraints, typical centralized approaches to
C-SLAM~\cite{anderssonCSAMMultiRobotSLAM2008,schmuckCCMSLAMRobustEfficient2019},
in which robots send measurements to a central computing node that computes and
shares back the merged global map and localization estimates, are highly
vulnerable to network disconnections or the outright failure of the central
computer.

To mitigate these risks, decentralized
techniques that are purposely built to work with ad-hoc networking and withstand
prolonged disconnections between robots are preferable. In prior work, we introduced
Swarm-SLAM~\cite{lajoieSwarmSLAMSparseDecentralized2024}, a decentralized C-SLAM
framework satisfying those requirements. Building on this framework, we
conducted a series of field experiments, and collected a novel dataset, at the Canadian Space Agency (CSA) Mars
Yard~\cite{CanadianSpaceAgency2021}, a planetary analogue terrain designed to
simulate realistic planetary conditions. We use Swarm-SLAM as a case study to
evaluate the current performance of state-of-the-art decentralized C-SLAM. We
look in particular at the challenges related to limited and intermittent
inter-robot communication, as well as the difficulties posed by the terrain in
terms of vibrations, lack of distinctive features, and perceptual aliasing.

\subsection{CONTRIBUTIONS}
\label{subsec:contributions}

This paper is the culmination of extensive multi-robot experiments on the
planetary analogue environments shown in~\cref{fig:csa-field}. As a result, we
present the following contributions:
\begin{itemize} 
\item The design of a decentralized three-robot system connected through ad-hoc
  networking, and its deployment on a planetary analogue environment;
\item A novel dataset collected during our experiments that includes LiDAR, IMU,
  and, unlike prior works, peer-to-peer inter-robot communication throughput and
  latency estimates that are valuable for evaluating the communication
  consumption of C-SLAM or other multi-robot algorithms. Our dataset\footnote{\href{https://ieee-dataport.org/documents/collaborative-simultaneous-localization-and-mapping-dataset-mars-analogue-terrain-inter}{https://ieee-dataport.org/documents/collaborative-simultaneous-localization-and-mapping-dataset-mars-analogue-terrain-inter}} is
  available on IEEE DataPort~\cite{lajoieCollaborativeSimultaneousLocalization2024};
\item A thorough analysis of the accuracy and efficiency of decentralized
  C-SLAM, exposing limitations of current approaches and open challenges.
\end{itemize} 
We believe that the challenges and lessons learned from our experiments will be
valuable for both the space robotics and C-SLAM research communities.

The remainder of this paper is divided as follows: \cref{sec:related_work}
presents background knowledge and related work on decentralized C-SLAM,
inter-robot networking, and localization challenges in space analogue
environments; \cref{sec:experimental_setup} describes our experimental setup
comprising the robots and sensors used and key characteristics of the terrain on
which they were deployed; \cref{sec:decentralized-cslam-results} presents the
accuracy results of our Swarm-SLAM decentralized C-SLAM algorithm and
localization challenges inherent to planetary analogue terrains;
\cref{sec:resource_efficiency} discusses the calibration trade-offs between
accuracy and resource efficiency in terms of communication and computing;
finally, \cref{sec:conclusion} offer some insights gained during our experiments
and open challenges that we identify for the future of C-SLAM for space
robotics.

\section{BACKGROUND AND RELATED WORK}
\label{sec:related_work}

\subsection{DECENTRALIZED COLLABORATIVE SIMULTANEOUS LOCALIZATION AND MAPPING}

\subsubsection{Centralized vs Decentralized}
C-SLAM systems are typically categorized into centralized or decentralized
solutions. Centralized C-SLAM relies on a main server or base station that
gathers mapping data from all participating robots and computes a common global
localization and mapping estimate for the whole team. This setup requires robots
to maintain a stable, continuous connection to the base station, leading to
significant challenges in scalability due to potential communication
bottlenecks. It is also vulnerable to failures in the central computing node~\cite{prorokRobustnessTaxonomyApproaches2021}.
Such stringent connectivity and reliability requirements can be impractical,
particularly in planetary environments without pre-existing networking
architectures.

In contrast, decentralized C-SLAM systems operate without a centralized server,
instead utilizing occasional, peer-to-peer communication among robots. This
approach is advantageous in environments where constant connectivity is
unreliable or impossible. However, decentralized systems face their own set of
challenges, as they are limited by the robots’ individual computing and
communication resources. They also demand more complex strategies for data
handling and coordination to ensure accurate and consistent SLAM results across
the team of robots~\cite{lajoieCollaborativeSimultaneousLocalization2022}.

Both centralized and decentralized C-SLAM systems share a similar architecture
with single-robot SLAM systems, consisting of two key components: the front-end
and the back-end. The front-end is responsible for tasks such as feature
extraction and data association, which involve identifying and matching
environmental features to aid in robot relocalization. The back-end, on the
other hand, is dedicated to state estimation, determining the robots' poses
(i.e. their positions and orientations) relative to the constructed map of the
environment~\cite{cadenaPresentFutureSimultaneous2016}. In collaborative state
estimation, the maps and poses of the robots are integrated into a common frame
of reference, allowing for consistent and unified positioning across all
robots~\cite{saeediMultipleRobotSimultaneousLocalization2016}.

\subsubsection{Front-End}

A major challenge in the C-SLAM front-end is efficiently detecting and computing
inter-robot loop closures. These loop closures occur when two or more robots
identify common landmarks or locations they have previously explored. Such
shared features act as connection points that allow the integration of local
maps from individual robots into a unified global reference frame. Because
transmitting entire maps can be prohibitively expensive in terms of
communication costs, inter-robot loop closure detection is typically handled in
a two-step
process~\cite{cieslewskiDataEfficientDecentralizedVisual2018,lajoieDOORSLAMDistributedOnline2020}.

In the first step, robots exchange compact descriptors, which are simplified
representations of their data, such as image descriptors (e.g.,
CosPlace~\cite{bertonRethinkingVisualGeoLocalization2022}) or LiDAR scan
descriptors (e.g., ScanContext~\cite{kimScanContextEgocentric2018}). These
descriptors enable place recognition by calculating similarity scores to
identify overlaps in the environments mapped by different robots. High
similarity scores suggest potential loop closure candidates, indicating that the
robots might have traversed the same place.

The second step focuses on these identified candidates. For each candidates with
high similarity, more detailed and computationally expensive descriptors, like
3D keypoints or full scans, are exchanged. These are used to perform precise
geometric registration between the corresponding data from different robots,
establishing accurate positional and rotational links between them. The
resulting pose measurements, called loop closures, are then integrated into the
robots’ pose graphs, to merge the maps and enhance the state estimation
accuracy.

In the case of LiDAR scans, since point clouds often include noise and outliers,
robust methods like TEASER++~\cite{yangTEASERFastCertifiable2021} are employed
to ensure accurate registration without needing an initial pose guess. This
capability is especially valuable in C-SLAM, as the robots generally do not know
their relative positions or trajectories prior to the first map merging.

\subsubsection{Back-End}

The C-SLAM back-end is tasked with estimating the most likely robot poses and
maps from the noisy data collected by the robots. To reduce the computational
cost of large-scale SLAM problems, which is especially challenging for
multi-robot mapping, most approaches utilize some form of pose graph
optimization. This method marginalizes environmental features into inter-pose
measurements, solving the optimization problem by focusing only on the poses.
While single-robot solvers can be directly applied to multi-robot scenarios, as
demonstrated in our approach~\cite{lajoieSwarmSLAMSparseDecentralized2024},
several distributed multi-robot solvers have been
developed~\cite{choudharyDistributedMappingPrivacy2017,tianDistributedCertifiablyCorrect2021,muraiRobotWebDistributed2024}.
These distributed approaches improve computational scalability by distributing the workload among connected robots. However, they require extensive bookkeeping, a large number of iterations during optimization, such that network delays can diminish their computational benefits. To mitigate these challenges, Fan and Murphey~\cite{fanMajorizationMinimizationMethods2024} propose a majorization minimization method for distributed pose graph optimization, which accelerates convergence with theoretical guarantees. Meanwhile, McGann and Kaess~\cite{mcgannIMESAIncrementalDistributed2024b} introduce an incremental distributed optimization approach that efficiently estimates robot states with only sparse pairwise communication. In contrast, our method streamlines optimization by dynamically electing a single robot to perform the computations during each rendezvous.

A common challenge in SLAM systems, including C-SLAM, is the occurrence of erroneous measurements due to perceptual aliasing~\cite{lajoieModelingPerceptualAliasing2019}. Perceptual aliasing arises when distinct locations in an environment appear overly similar and lack distinguishing features, leading to incorrect data associations during place recognition.
In C-SLAM systems, this challenge has been well-documented across various environments. For example, Ebadi et al.\cite{ebadiLAMPLargeScaleAutonomous2020} highlight the impact of perceptual aliasing in subterranean settings, while Tian et al.\cite{tianSearchRescueForest2020} report frequent occurrences in forested environments.
To mitigate this issue, Mangelson et
al.\cite{mangelsonPairwiseConsistentMeasurement2018} introduced Pairwise
Consistency Maximization (PCM), which enhances robustness by identifying the
largest set of consistent inter-robot measurements. In another line of work,
Yang et al.\cite{yangGraduatedNonConvexityRobust2020} proposed the Graduated
Non-Convexity (GNC) algorithm, a flexible and robust tool for various
optimization tasks. GNC is employed by the elected robots in Swarm-SLAM for pose
graph optimization. Tian et al.~\cite{tianKimeraMultiRobustDistributed2022} have
extended GNC for use in distributed implementations.

Although visual loop closure detection has advanced significantly~\cite{bertonRethinkingVisualGeoLocalization2022,sarlinSuperGlueLearningFeature2020}, vision-based methods face major challenges in planetary analogue environments. Changes in appearance due to varying viewpoints—exacerbated by the limited field of view of cameras—hinder reliable detection. Additionally, variations in illumination and the presence of dust can degrade image quality, while the scarcity of unique visual features in these environments often leads to perceptual aliasing, complicating data association~\cite{ebadiRoverLocalizationMars2020}.
In contrast, LiDAR is well-suited for mapping in visually degraded conditions, as it operates independently of ambient light and provides precise 360° range measurements. To mitigate the limitations of both sensors, cameras are frequently integrated with LiDAR, leveraging their complementary properties to enhance loop closure detection~\cite{newmanNavigatingRecognizingDescribing2009,shanLVISAMTightlycoupledLidarVisualInertial2021}.

Recently, numerous C-SLAM frameworks have emerged, each contributing to
advancements in distributed mapping.
DSLAM~\cite{cieslewskiDataEfficientDecentralizedVisual2018} was a pioneering
system that used CNN-based image descriptors for place recognition and
distributed pose graph optimization.
DOOR-SLAM~\cite{lajoieDOORSLAMDistributedOnline2020} further developed these ideas by
integrating PCM for outlier rejection and adapting the system to handle
intermittent inter-robot communication, eliminating the need for full
connectivity between robots. DiSCo-SLAM~\cite{huangDiSCoSLAMDistributedScan2022}
expanded these concepts to LiDAR-based mapping, using ScanContext
descriptors~\cite{kimScanContextEgocentric2018} for place recognition.
Kimera-Multi~\cite{tianKimeraMultiRobustDistributed2022} combines classical
approaches to place recognition and registration with a distributed version of
GNC. 
More recently, SlideSLAM~\cite{liuSlideSLAMSparseLightweight2024} introduced a decentralized metric-semantic SLAM framework that enables heterogeneous robot teams to collaboratively build and merge object-based maps. 

Centralized C-SLAM systems have also advanced considerably.
COVINS~\cite{schmuckCOVINSVisualInertialSLAM2021} optimizes visual-inertial SLAM
by reducing computational load through the elimination of redundant keyframes.
LAMP 2.0~\cite{changLAMP20Robust2022} employs a Graph Neural Network-based
prioritization mechanism to evaluate inter-robot loop closure candidates,
predicting the optimization outcomes and selecting the most promising candidates
for further processing. Maplab 2.0~\cite{cramariucMaplab20Modular2023} is
designed to support varying sensor modalities and configurations, enabling
flexible multi-robot mapping.

Additionally, related to our work, the ARCHES project~\cite{schusterARCHESSpaceAnalogueDemonstration2020} explores heterogeneous robotic teams for collaborative navigation and sampling, demonstrating multi-robot 6D SLAM~\cite{schusterDistributedStereoVisionbased2019a} on a planetary analogue environment.

\ExecuteMetaData[explanation_figures.tex]{overview}

Swarm-SLAM~\cite{lajoieSwarmSLAMSparseDecentralized2024}, our recently proposed
system described in \cref{fig:swarm-slam-overview}, builds on these advancements
by introducing a novel sparse inter-robot loop closure prioritization technique
to reduce communication overhead. It uses ROS~2 and includes a neighbor
management system that integrates smoothly with ad-hoc networking, enhancing
C-SLAM’s adaptability to intermittent communication scenarios. For a
comprehensive review of C-SLAM technologies, we refer the readers
to~\cite{lajoieCollaborativeSimultaneousLocalization2022}.

\subsection{AD-HOC INTER-ROBOT COMMUNICATION}

Ad hoc networks play a crucial multi-modal role in enabling multi-robot mapping, allowing
robots to communicate directly with one another without relying on a
pre-existing infrastructure. In the early exploration of ad-hoc inter-robot
communication for collaborative mapping, Sheng et
al.~\cite{shengMinimizingDataExchange2005} proposed a 2D grid-based approach
that minimizes data exchange by leveraging known relative poses between robots.
More recently, Varadharajan et al.\cite{varadharajanSOULDataSharing2020}
addressed the broader challenge of efficiently sharing large volumes of data,
such as maps, within distributed robot networks by introducing a peer-to-peer
data sharing system specifically designed for high data loads. To ensure
reliable inter-robot communication, it is crucial to consider the robot
topology, as it directly influences the available communication paths between
connected
agents~\cite{ghediniEfficientAdaptiveAdhoc2018,siligardiRobustAreaCoverage2019}.
For example, Varadharajan et al.~\cite{varadharajanSwarmRelaysDistributed2020}
proposed a fully decentralized connectivity algorithm robust against individual
robot failures. This approach allows robots to autonomously adjust their
positions to maintain network connectivity with a ground station, ensuring
stable communication despite dynamic conditions and potential disconnections.
In the specific context of C-SLAM, Giamou et al.\cite{giamouTalkResourceEfficientlyMe2018} and Tian et al.\cite{tianOptimalBudgetedData2018} acknowledge the communication bottleneck issue and developed near-optimal data-sharing strategies to avoid duplication during rendezvous between multiple robots.

While some C-SLAM systems, like those described
in~\cite{cieslewskiDataEfficientDecentralizedVisual2018,schmuckCOVINSVisualInertialSLAM2021}
require fully connected networks, recent approaches have been designed to be
resilient to disconnections. For instance, our prior
work~\cite{lajoieDOORSLAMDistributedOnline2020} and Tian et
al.~\cite{tianResilientDistributedMultiRobot2023} propose C-SLAM frameworks that
can withstand intermittent communication losses. More
recently~\cite{lajoieSwarmSLAMSparseDecentralized2024}, we successfully deployed
C-SLAM with ad-hoc networking in real-world settings, demonstrating the
practical feasibility of decentralized multi-robot mapping in challenging
environments without the need for complex continuous connectivity maintenance.

\ExecuteMetaData[explanation_figures.tex]{datasettable}

\ExecuteMetaData[explanation_figures.tex]{field-examples}

\subsection{MULTI-ROBOT MAPPING DATASETS}

Existing datasets in C-SLAM typically fall short of capturing realistic
multi-robot scenarios, as they often involve only one robot at a time, resulting
in the absence of dynamic objects, and systematically lack inter-robot network
conditions estimates. This represents a significant gap in the literature,
particularly regarding experimental data from planetary analogue environments
where inter-robot communication is intermittent due to large distances and
obstacles that cause non-line-of-sight conditions between robots.

One of the early efforts in multi-robot datasets was the UTIAS dataset by Leung
et al.~\cite{leungUTIASMultirobotCooperative2011}, which involved five robots
operating within a single indoor room. This dataset set the groundwork for
collaborative SLAM but was limited to a static, confined environment. More
recently, Dubois et al.~\cite{duboisAirMuseumHeterogeneousMultirobot2020}
proposed incorporating both ground and aerial robots in indoor settings, using
stereo cameras. Collected during larger scale outdoor environment, Zhu et
al.~\cite{zhuGRACOMultimodalDataset2023} introduce GrAco a multimodal dataset
featuring ground and aerial LiDAR and stereo sequences captured on a college
campus. This dataset offered more diverse environmental settings but remained
limited to structured outdoor spaces. A notable step forward in scaling and
realism came from Tian et al.~\cite{tianResilientDistributedMultiRobot2023}, who
developed an online dataset featuring up to eight robots equipped with cameras
and LiDAR, operating in large-scale, indoor and outdoor environments with
human-made structures. Feng et al.~\cite{fengS3EMultiRobotMultimodal2024}
introduced the multimodal S3E dataset, which specifically targets C-SLAM scenarios with
three synchronized robots in multiple indoor and outdoor environments.
Park et al.~\cite{parkBenchmarkDatasetCollaborative2024} introduce a novel C-SLAM simulated dataset designed for service environments, where the presence of dynamic objects poses significant challenges.

Most relevant to our work, Zhao et al.~\cite{zhaoSubTMRSDatasetPushing2024}
addressed some of these limitations with the SubT-MRS dataset, which includes
diverse robots operating in various environments, including challenging,
degraded conditions similar to the planetary analogue field used in our
experiments. This dataset significantly contributes to the field by simulating
more realistic conditions for multi-robot systems. Our dataset aims to further
advance robustness and resilience in degraded environments, while also providing
data on real inter-robot communication capacity in the field. To that end, our
dataset, described in~\cref{tab:dataset_table}, includes periodic pairwise
latency and throughput estimates.

\section{EXPERIMENTAL SETUP}
\label{sec:experimental_setup}

Our experiments were conducted in a planetary analogue field designed to simulate the
challenging conditions of planetary surfaces. The terrain included a mix of
sand, various types of rocks, slopes, and uneven ground, closely mimicking the
environment that robots would encounter on actual space missions
(see~\cref{fig:terrain-pictures}). These features pose significant challenges
for robot mobility, perception, and communication, making this setup ideal for
testing decentralized C-SLAM algorithms in realistic conditions. 
It is important to note that, since the environment is on Earth, it is not fully realistic, as there is vegetation in the distance and human operators moving around the area.

The experiments involved three robots exploring the field simultaneously, each
remotely controlled by human operators. This setup allowed for testing the
robots' ability to maintain ad hoc communication and correctly localize
themselves in a dynamic and degraded environment. As illustrated
in~\cref{fig:pose-graph-viz}, the robots followed roughly similar trajectories,
but each in different order and directions. This approach ensured that each
robot covered nearly the entire field, maximizing overlap between the maps, and
thus producing as many loop closing matches as possible for our system analysis
in~\cref{sec:resource_efficiency}.

\subsection{ROBOT DESIGN}
Our experiments were conducted using three robots: one AgileX Scout 2.0 Rover and two AgileX
Bunkers, shown in~\cref{fig:robots}. All
robots were mounted with NVIDIA Jetson Xavier (32GB) for data processing, GL
iNet AX1800 router for wireless network, Ouster LiDAR OS0, Vector IMU VN-100 and
U-Blox ZED F9 modules for GPS positioning (F9P for bunkers and F9R for the
rover). Our Swarm-SLAM system was configured with LiDAR as the primary sensor
for environment mapping and localization. For odometry, we used the LIO-SAM
algorithm~\cite{shanLIOSAMTightlycoupledLidar2020}, which integrates LiDAR and
inertial data to provide accurate and robust local pose estimation in real-time.
For place recognition, we employed
ScanContext~\cite{kimScanContextEgocentric2018}, a method that generates compact
descriptors of LiDAR scans for recognizing previously visited locations. To
determine similarities between locations, we used cosine similarity to compare
the ScanContext descriptors. To ensure robust 3D registration of the point
clouds matched with ScanContext, we integrated
TEASER++~\cite{yangTEASERFastCertifiable2021}, a state-of-the-art algorithm
designed for fast and certifiable registration of point clouds.
To evaluate Swarm-SLAM's odometry estimates, the robots were
equipped with U-Blox ZED GPS modules.
The modules on the robots were programmed to correct their GPS estimates by receiving RTCM correction data (i.e. Radio Technical Commission for Maritime Services messages) from our GPS station. 
Our GPS RTK (i.e. Real-Time Kinematic positioning) station comprises of a laptop connected to our ad-hoc network, and a U-Blox ZED GPS module.
First the module is configured to perform the \textit{survey-in} procedure, where the module calibrates its position with an accuracy of up to 20cm.  
After reaching the target accuracy, our GPS station starts a NTRIP (Networked Transport of RTCM via Internet Protocol) server to broadcast the RTCM correction data on the network. 
As a result, our system provides GPS estimates at 1 Hz, though occasional delays occur due to communication latency in transmitting correction messages from the GPS station to the robots. Meanwhile, the U-Blox driver publishes the most recent estimate at approximately 10 Hz.

\ExecuteMetaData[explanation_figures.tex]{robot}

\subsection{AD-HOC NETWORKING}
\begin{figure}
  \centering
  \includegraphics[width=\columnwidth]{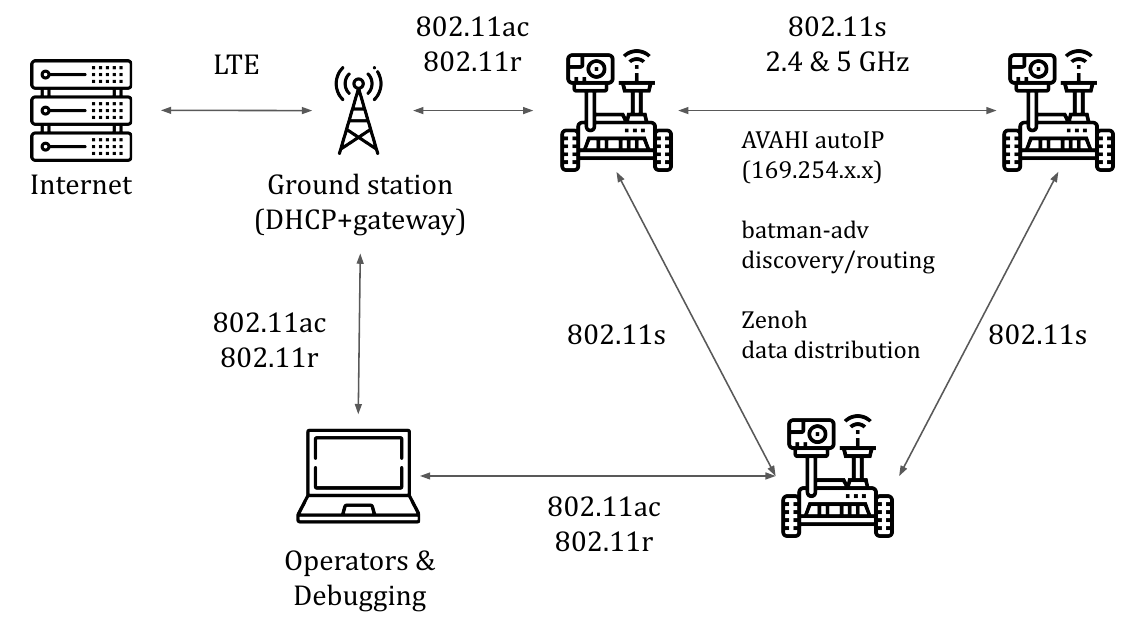} \caption{Network
    architecture: Robots are interconnected via 802.11s on bonded 2.4GHz and
    5GHz interfaces, with routing provided by batman-adv and data distribution
    managed by Zenoh.}
  \label{fig:network}
\end{figure}

We developed a Mobile Ad-Hoc Network (MANET) using a custom network stack
implemented on GL iNet AX1800 routers, which run
OpenWRT~\cite{fainelli2008openwrt}. Each robot, along with the ground station,
was equipped with a router responsible for managing both internal communications
among the robot’s hardware components and external communications with other
robots via the MANET. The inter-robot links were based on IEEE 802.11s, combined
with batman-adv, a dynamic link-state routing protocol operating at the data
link layer (Layer 2). Batman-adv continuously broadcasts network updates,
maintaining a routing table that ensures seamless communication between nodes
(i.e. robots) throughout the deployment. The overall communication architecture
is illustrated in \cref{fig:network}.

The MANET backbone relies on IEEE 802.11s, configured without frame replication
to optimize bandwidth usage and allow batman-adv to provide adaptive routing.
Key parameters were adjusted to support fast adaptation to network changes (root
mode, active path timeouts, etc.), allowing for rapid disconnection and
reconnection under low signal conditions. The 2.4GHz and 5GHz radios were bonded
into a shared interface managed by batman-adv, enabling dynamic frequency
switching. This setup leverages the range of 2.4GHz, which is more susceptible
to interference but offers wider coverage, and the higher bandwidth of 5GHz,
which is better suited for shorter-range, high-data-rate communications. This
dual-frequency capability allows the network to automatically choose the optimal
frequency and routing path, supporting both direct and multi-hop communication.
We also consider an optional 900MHz channel for specific use cases requiring
long-range communication (several kilometers), which was not needed given the
 size of the CSA Mars Yard. In fact, we used the standard
antennas present on the AX1800 routers to have a more challenging communication
environment with ranges up to 40 meters, whereas high-gain antennas or more
powerful routers could have provided communication range up to hundreds of
meters and covered the entire terrain.

To ensure connectivity with the operator computers, the bonded interface was
bridged with a 802.11ac link in station mode and the robots' local Ethernet
network. In other words, each robot acts as a wireless access point, and all the
on-board computers share the same subnet. The network incorporates 802.11r for
fast transitions, enabling operator laptops to seamlessly connect and switch
between access points, whether on the ground station or directly to robots. All
robots were configured to operate on a unified subnet, while VLANs (i.e. Virtual Local Area Networks) were used to
segregate local components outside the bridge, reducing interference and
preventing flooding (e.g., isolating LiDAR data streams from other local network
traffic). All computers were set up to use AVAHI AutoIP (i.e. automatic IP address configuration) for service and name
discovery under the .local domain. All data distribution between robots is
provided by Zenoh~\cite{corsaro2023zenoh}, which offers a minimal overhead publish-subscribe
communication API.

The ground station was configured similarly to the robots but had an additional
gateway for Internet access. This gateway facilitated software updates for the
robots and synchronized their system clocks, ensuring consistent timing across
the network. Furthermore, the ground station acted as a control hub, enabling
operators to monitor robot status, trigger behaviors, and manage overall mission
coordination within the network’s communication range.


\subsection{PEER-TO-PEER BANDWIDTH ESTIMATION}

To assess the communication performance between the robots, we conducted pairwise peer-to-peer communication estimations at 1-second intervals throughout our data collection experiments. The cumulative throughput was calculated by summing throughput over consecutive 1-second intervals.
 We measured throughput using
iperf~\cite{dunganIPerfTCPUDP}, a tool designed for active measurements of the
maximum achievable bandwidth. For latency measurements, we employed
fping~\cite{schemersFPing}, a program that sends ICMP (i.e. Internet Control Message Protocol) echo probes to calculate
round-trip times between the robots. This approach enabled us to monitor the
latency of communications dynamically as the robots moved through the Mars
analogue field, reflecting the impact of varying distances, obstacles, and
network topology changes on communication delays. To ensure that our estimates
were as accurate and reflective of real-world conditions as possible, we
minimized additional network traffic by not running any other software that
required data transmission. 
Thus, we ran our C-SLAM benchmark in post-processing to assess how well the approach aligns with real-world limitations.
On top of serving as a benchmark for inter-robot communication, we believe our dataset provides valuable insights into network performance under dynamic conditions. It can assist researchers in optimizing data transmission strategies, evaluating adaptive bandwidth allocation methods, and developing robust communication protocols for multi-robot systems operating in challenging environments.


\section{DECENTRALIZED COLLABORATIVE SLAM}
\label{sec:decentralized-cslam-results}

\ExecuteMetaData[experimentation_figures.tex]{swarm-slam-solution}

\ExecuteMetaData[experimentation_figures.tex]{swarm-slam-solution-indv}

We used Swarm-SLAM to process the entire data sequences from the robots,
generating the multi-robot pose graph solution illustrated
in~\cref{fig:pose-graph-viz}. The GPS ground truth data is also displayed for
comparison. To quantify the accuracy of the SLAM estimates, we employed the evo
software~\cite{gruppEvoPythonPackage2017} to compute error metrics.
We aligned the pose estimates with the GPS data points with the closest timestamps.
The results
showed an Average Translation Error (ATE) of 3.74 $\pm$1.63 meters between the
estimated poses and the ground truth. 
For conciseness, in this paper, we report the ATE as $\mu \pm \sigma$, where $\mu$ is the mean error and $\sigma$ is the standard deviation.
The trajectory estimates are jointly aligned with the GPS ground truth using the Umeyama method to evaluate the accuracy of inter-robot localization.
We also provide the individual estimates in~\cref{fig:swarm-slam-solution-indv}.
For reproducibility, we provide our solution along with the dataset, as well as processing and evaluation scripts.
Interestingly, while reasonable, our
results are far from perfect, suggesting that future research could leverage our
dataset to develop new techniques that improve localization and mapping accuracy
in C-SLAM within this type of difficult environment.

\ExecuteMetaData[experimentation_figures.tex]{error-distribution}

A detailed breakdown of the error distribution per robot is presented
in~\cref{fig:error-distribution}. Notably, the error for Robot 1 is
significantly lower than that of Robots 2 and 3. We hypothesize that this
variation is due to differences in robot design. While Robot 1 features large
wheels and suspension, which is well suited for navigation on sandy terrain with
numerous small rocks, Robots 2 and 3 are equipped with tracks, which are less
effective in these conditions. To verify this hypothesis, we analyzed the linear
acceleration data from the IMUs embedded in the 3D LiDAR sensors of each robot,
shown in~\cref{fig:lin-acc}. This revealed that Robots 2 and 3 experienced
significantly higher vibration levels compared to Robot 1, indicating that the
wheeled configuration of Robot 1 is more compatible with the challenging
terrain.

\ExecuteMetaData[experimentation_figures.tex]{lin-acc}

These increased vibrations in Robots 2 and 3 likely led to reduced LiDAR
odometry accuracy as it may cause the loss of points and hinder data
association~\cite{schlagerAutomotiveLidarVibration2022}. To confirm this, we
measured the ATE for the odometry of each robot individually: Robot 1 exhibited
an odometry ATE of 2.45$\pm$1.14 meters, while Robot 2 and Robot 3 had higher
ATEs of 4.29 $\pm$1.76 meters and 3.61 $\pm$1.72 meters, respectively.

Despite the wheeled robot’s better performance in terms of reduced error, it
faced its own challenges. Robot 1, with its large wheels, was more susceptible
to becoming stuck, especially in wet sand, whereas the track-equipped Robots 2
and 3 demonstrated superior traction and reliability.

\section{RESOURCE EFFICIENCY}
\label{sec:resource_efficiency}

This section delves into the resource efficiency of our decentralized C-SLAM
solution when deployed in planetary analogue environments. Decentralized C-SLAM
systems must operate within the constraints of available computing power,
memory, and inter-robot communication resources, requiring strategic trade-offs
to ensure effective performance under these limited conditions.

In \cref{subsec:inter-robot-communication}, we analyze the inter-robot
communication metrics gathered during our experiments and compare them to the
default requirements of our C-SLAM approach. This analysis provides a critical
evaluation of the realism and feasibility of our solution in practical
scenarios, assessing whether the system’s communication demands are compatible
with the actual network conditions experienced in the field.

\cref{subsec:calibration} addresses the calibration of key parameters to tune
the C-SLAM system for real-world deployments. We explore the trade-offs between
map accuracy and available communication bandwidth, illustrating how adjusting
these parameters can influence the system’s overall performance. Clear
understanding of these trade-offs is essential, as it enables more informed
decisions regarding the suitability of the approach for different mission
scenarios.

Moreover, well-defined trade-offs significantly enhance the tunability of the
system. A solution that is easier to calibrate not only streamlines deployment
but also increases the likelihood of adoption, particularly by users who are not
SLAM experts. By simplifying the tuning process, the technology becomes more
accessible and adaptable, making it a practical choice for a wider range of
applications and user groups.

\subsection{AD-HOC INTER-ROBOT COMMUNICATION}
\label{subsec:inter-robot-communication}

\ExecuteMetaData[experimentation_figures.tex]{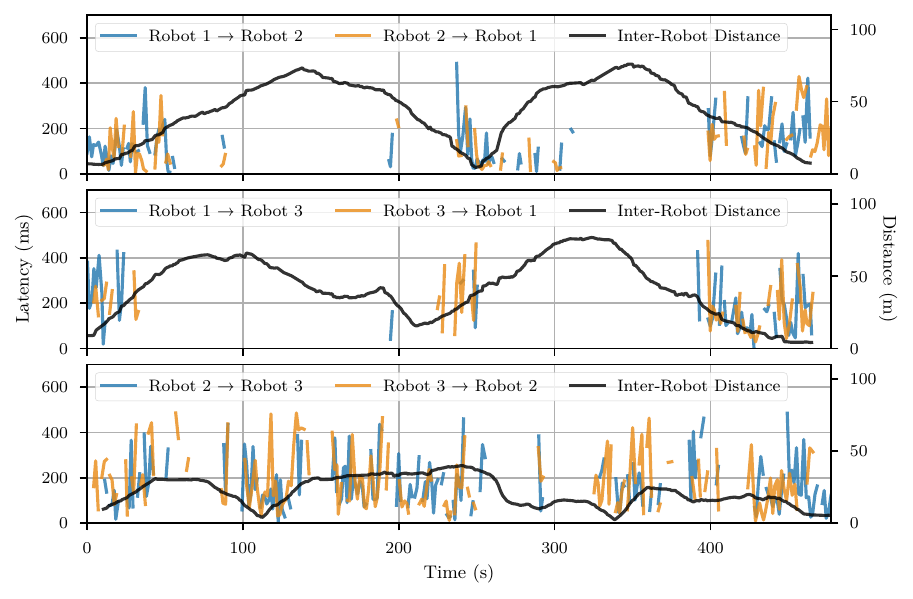}

We analyzed the available pairwise peer-to-peer inter-robot communication
bandwidth during our field mission, as peer-to-peer communication is the
backbone of scalable and resilient multi-robot operations. A system that relies
primarily on local inter-agent data transmission can better scale to large
groups of robots because it avoids the need for a central communication node,
which can become a bottleneck or critical point of
failure~\cite{kegeleirsSwarmSLAMChallenges2021}. To demonstrate the capabilities
of our system, we present latency measurements in both directions for each of
the three robot pairs in \cref{fig:latency}. The observed latencies range from
100ms to 400ms, which, even without considering computation time, imposes
significant constraints on real-time C-SLAM deployment.

To address this challenge, our approach maintains a real-time, local
single-robot SLAM estimates, which are periodically updated and corrected using
the multi-robot estimates that incorporate the multi-robot pose graph and
inter-robot loop closures. Additionally, \cref{fig:latency} shows the distance
between robots at each timestep, calculated using GPS data. The comparison
between latency and inter-robot distance indicates that our networking setup
tends to lose connectivity when robots are approximately 40 meters or more
apart. This finding highlights the need for approaches that can handle
disconnections and effectively recompute a consistent map when robots reconnect.

\ExecuteMetaData[experimentation_figures.tex]{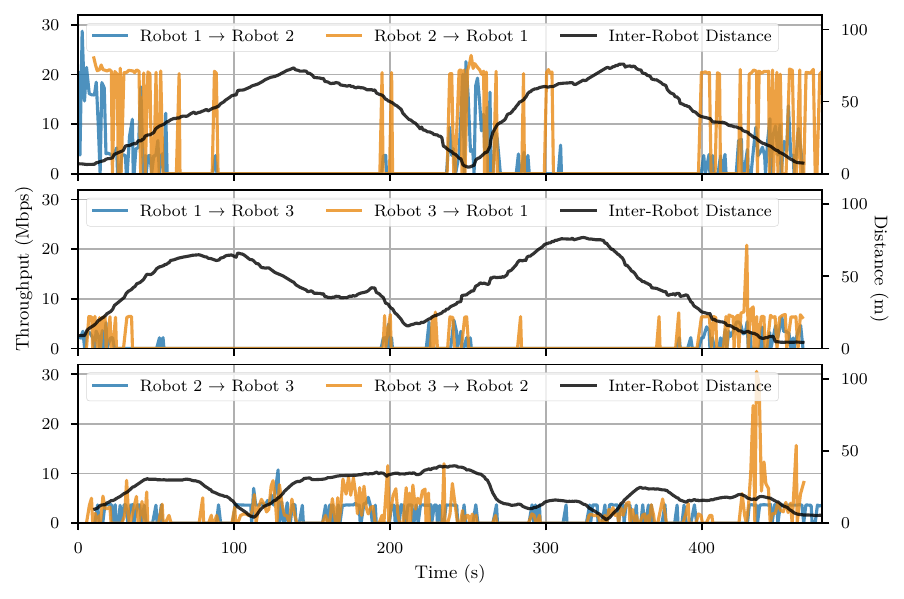}

In \cref{fig:throughput}, we also present the throughput between robots in both
directions for each pair, with values ranging from approximately 5 to 20 Mbps
when connected. Comparing throughput with inter-robot distances confirms the
consistency of our latency estimates with the throughput data. 

\ExecuteMetaData[experimentation_figures.tex]{comm-comp}
Using these
throughput estimates, \cref{fig:comm-comp} illustrates the accumulated
communication throughput over time across all robot pairs, plotting the
cumulative available communication bandwidth in megabytes at each timestep.
We compared this field-measured bandwidth with the unconstrained bandwidth usage
of Swarm-SLAM. To accurately measure the unconstrained bandwidth usage, or
maximum bandwidth consumption, we ran three agents in parallel on a single
machine, each processing sensor data from one robot in our dataset. We mesured
all data transmission through the ROS~2 nodes of different agents,
distinguishing between back-end and front-end processes to better identify their
relative bandwidth requirements. Our findings indicate that front-end processes
dominate the overall communication load. A key insight from \cref{fig:comm-comp}
is that while available bandwidth is initially sufficient, communication demands
rapidly increase, eventually exceeding the available capacity. This is
attributed to the initially small individual robot maps with limited overlap,
resulting in minimal need for resource-intensive 3D registration. However, as
each robot explores and expands its map, the level of overlap with other robots
increase. Subsequently, the number of successful place recognition matches
grows, leading to a rise in communication requirements. Interestingly, this
shows that although increased map overlap will ultimately enhance map merging
accuracy, it also demands more communication and computation to be effectively
processed. Therefore, our experiments represent a challenging scenario in terms
of inter-robot bandwidth due to the substantial overlap between robot
trajectories and maps, making this a valuable case study for understanding the
communication limits of C-SLAM systems.

It is important to note that the results in \cref{fig:comm-comp} were obtained
using default parameters: a ScanContext cosine similarity threshold of 0.7 and a
minimum of 80 inlier points for registration. As will be discussed in
\cref{subsec:calibration}, these settings generate numerous loop closure
candidates, including incorrect ones, and may not be the most
communication-efficient. We will explore how strategic tuning of these
parameters can enhance communication performance without compromising the
accuracy of the C-SLAM solution.

\subsection{IMPACTS OF C-SLAM CALIBRATION}
\label{subsec:calibration}

Calibrating key parameters of decentralized C-SLAM systems is crucial for
meeting communication constraints, which is especially important in
resource-limited environments like planetary analogue settings. 

\subsubsection{Communication Budget}
In Swarm-SLAM~\cite{lajoieSwarmSLAMSparseDecentralized2024}, we introduced a
communication budget, defined as the number of inter-robot loop closure matches
selected from all candidate matches identified through place recognition. This
budget uses a spectral sparsification approach to prioritize candidate matches
before they are send to the, more communication and computation-intensive, 3D
registration step, as shown in~\cref{fig:swarm-slam-overview}.

\ExecuteMetaData[experimentation_figures.tex]{comm-budget}

The prioritization process focuses on selecting matches that are most likely to
improve the accuracy of the multi-robot pose graph, making it particularly
valuable when high map overlap generates an excess of place recognition matches
beyond available resources. It is also useful when robots reconnect after
extended periods of disconnection, during which they accumulate a backlog of
place recognition matches that could take significant time and communication to
process entirely. By carefully prioritizing which matches to process, we can
improve the trade-off between communication and accuracy, allowing a small
number of well-chosen loop closures to closely approximate the optimal C-SLAM
solution.

Therefore, the communication budget directly controls how much data Swarm-SLAM
transmits at each timestep. In \cref{fig:comm-budget}, we show the optimal match
selection budget at each timestep versus the cumulative throughput. In practice,
this budget is often set to a fixed value, but our results suggest that adapting
the budget dynamically could better utilize available bandwidth. However,
evaluating throughput online during experiments poses a challenge as most
estimation techniques require sending large volumes of data to test the limits
of the network, which could interfere with ongoing communication.

\ExecuteMetaData[experimentation_figures.tex]{sim-vs-loops}

\subsubsection{Place Recognition}
In \cref{fig:sim-vs-loops}, we explore the relationship between the number of
loop closures (y-axis) and the place recognition similarity threshold (x-axis).
We categorize loop closures into three groups: correct loop closures (in green)
with translation errors below the average error of the optimized multi-robot
pose graph (i.e. 3.74 meters), less accurate or incorrect loop closures (in yellow) with errors
above the average, and failed matches (in red) where high descriptor similarity
did not result in successful registration due to insufficient inlier points. The
plot demonstrates that lower, less conservative, similarity thresholds result in
more loop closures, but many of these are incorrect or failed registrations.
Conversely, increasing the threshold reduces incorrect matches but also
significantly decreases the number of correct ones within a certain range (0.7
to 0.85), revealing a trade-off between conservativeness and loop closure
quantity. This gap indicates that current similarity measures, such as those
used in ScanContext, cannot perfectly predict the quality of loop closures
post-registration.

To isolate the effect of the similarity threshold, we used a very loose
threshold of only 10 inlier points for subsequent 3D registration. Although
robust pose graph optimizers like GNC~\cite{yangGraduatedNonConvexityRobust2020}
can tolerate some incorrect or outlier loop closures, they incur significant
computational costs, often requiring several seconds compared to milliseconds
for standard optimizers. Reducing the number of incorrect loop closures through
better front-end calibration could thus lead to notable efficiency gains in
back-end optimization.

In the same figure, the right y-axis shows the communication cost in terms of
KBytes per correct loop closure. Generally, more conservative similarity
thresholds lead to better communication cost due to fewer incorrect loop
closures. However, while conservative thresholds may work well in high-overlap
scenarios like our experiments, they risk missing inter-robot loop closures in
environments with less map overlap, where loop closures are more rare. Thus, in
low-overlap scenarios, it may be advisable to use less conservative thresholds
in order to perform map merging and achieve a C-SLAM solution.

Our analysis of the relationship between similarity thresholds and the balance of correct and incorrect loop closures extends beyond our technique. It is relevant to many C-SLAM frameworks~\cite{cieslewskiDataEfficientDecentralizedVisual2018,tianKimeraMultiRobustDistributed2022} that employ a two-stage loop detection approach—place recognition followed by registration—to improve efficiency.

\ExecuteMetaData[experimentation_figures.tex]{inliers-vs-loops}

\ExecuteMetaData[experimentation_figures.tex]{ambiguity}

\subsubsection{Registration}
In \cref{fig:inliers-vs-loops}, we examine the number of loop closures (correct
in green, incorrect in yellow) relative to the number of inlier points during
registration, using a very low similarity threshold of 0.1 to focus on the
number of inliers effect. The results show that, again, there is a trade-off
between setting more conservative thresholds and the total number of loop
closures. Unfortunately, this parameter alone is not a reliable predictor of
loop closure accuracy, as some loop closures with over 300 inliers were still
incorrect. Consequently, more conservative thresholds can worsen communication
efficiency, as they reduce correct loops without effectively filtering out
incorrect ones.

As shown in \cref{fig:sim-vs-loops,fig:inliers-vs-loops}, our experiments
indicate that planetary analogue environments are prone to place recognition
outliers and inaccurate 3D registrations. The flat terrain and lack of
distinctive features often cause different places to appear similar. In
\cref{fig:matchingambiguity}, we illustrate this phenomenon with two point
clouds with a high ScanContext similarity of 0.757 and a substantial number of
inliers (381), which, despite looking similar, represent distinct locations
17.28 meters apart. Importantly, as we have shown, these outliers not only
affect overall accuracy but also negatively impact the communication cost.

\section{CONCLUSIONS AND OPEN CHALLENGES}
\label{sec:conclusion}

In this paper, we presented a comprehensive evaluation of decentralized C-SLAM
in planetary analogue environments, addressing the unique challenges posed by
difficult terrain, limited resources, and the need for efficient communication
strategies. Our experiments highlighted several critical insights into deploying
decentralized C-SLAM in such challenging settings, where the terrain affects
robot mobility and sensor data quality due to vibrations and uneven surfaces.
These conditions underscore the need for robust, adaptable SLAM algorithms
capable of maintaining accuracy in uncontrolled environments.

One of the primary challenges identified is the constraint imposed by limited resources—namely, communication bandwidth, computational power, and memory. Effective operation in such environments demands careful and adaptive tuning of system parameters, with a significant emphasis on optimizing communication, which frequently emerges as the most limiting factor. Our findings show that the C-SLAM front-end consumes the bulk of the communication bandwidth, underscoring the need for future research to reduce its demands. 
Potential strategies include techniques such as advanced point cloud compression~\cite{houshiar3DPointCloud2015}, descriptor-based representations~\cite{han3DPointCloud2023}, and voxelization~\cite{xuVoxelbasedRepresentation3D2021} could be readily integrated into our approach. However, tuning lossy data compression in feature-sparse environments, such as the one in our dataset, presents challenges—important salient features could be inadvertently discarded, increasing the risk of perceptual aliasing.
Alternative approaches, such as semantic-level descriptors~\cite{changHydraMultiCollaborativeOnline2023}, could further reduce communication overhead but would require a fundamental system redesign. These methods also come with their own limitations, such as reliance on the presence of semantically rich features like man-made objects, which may not always be available in the environment.

Moreover, our study revealed a persistent trade-off between communication and
accuracy in current C-SLAM approaches. Enhancing this trade-off remains an open
challenge, with future research potentially focusing on improving accuracy
without proportionately increasing communication and computational demands. This
could be achieved by refining existing algorithms, exploring novel sensing and
mapping paradigms, or developing more efficient data fusion techniques.

As the number of robots in a collaborative SLAM system increases, so does the complexity of data exchange. More robots can generate additional inter-robot loop closures, which could enhance mapping accuracy. However, this also introduces significant bandwidth contention, as multiple robots attempt to share data within the same network. Our decentralized approach offers a promising pathway for scalability, as it reduces reliance on a central server and allows robots to make localized decisions. However, to maintain communication feasibility in large-scale swarms, hierarchical strategies will likely be necessary. For instance, clustering robots into subnetworks or prioritizing loop closure transmissions based on relevance and certainty could help mitigate bandwidth congestion while preserving accuracy gains.

To advance C-SLAM research, we provide a dataset collected during our field experiments, which includes inter-robot latency and throughput monitoring over our ad-hoc network. This dataset can be used to simulate or model realistic channel conditions. For example, one could integrate measured latency and bandwidth traces into a simulator or offline approach, gating data transmissions based on real bandwidth constraints. Additionally, we have included example scripts for communication modeling, particularly for analyzing latency and throughput as functions of inter-robot distance. We hope this dataset will support other research groups in designing novel communication-aware C-SLAM approaches.

Our Swarm-SLAM approach is intentionally designed to be general and applicable
across a wide range of scenarios, beyond just planetary analogue environments.
It relies on inexpensive onboard sensors and simple peer-to-peer communication
links, making it particularly suitable for early space missions where permanent
networking or localization infrastructure---such as satellites or base
stations---has yet to be established. However, the integration of infrastructure
like orbital satellites or base stations with long-range, high-power networking
and localization capabilities could significantly enhance SLAM performance by
providing external or global sensing for the entire group of robots.

We believe that, in the future, the most effective approaches for space
exploration will involve a hybrid strategy that fuses local sensing and
estimation with global sensing capabilities. This fusion would ensure safe and
reliable autonomy through local sensing and allow for decentralized inter-robot
mapping during periods of communication loss or base station outages, while
benefiting from external sensing and larger computing resources when available.
This balanced integration of local autonomy and global coordination is key to
overcome the unique challenges of operating in extraterrestrial environments.

Ultimately, advancing C-SLAM technology for space exploration will require
continuous refinement of these adaptive strategies to meet the evolving demands
of complex and resource-constrained environments. By enhancing our understanding
of the trade-offs between communication, computation, and accuracy, we can
better equip multi-robot systems to navigate and map new frontiers---whether on
Earth, the Moon, Mars, or beyond.

\bibliographystyle{IEEEtran}
\bibliography{references}

\begin{thebibliography}{10}
\providecommand{\url}[1]{#1}
\csname url@samestyle\endcsname
\providecommand{\newblock}{\relax}
\providecommand{\bibinfo}[2]{#2}
\providecommand{\BIBentrySTDinterwordspacing}{\spaceskip=0pt\relax}
\providecommand{\BIBentryALTinterwordstretchfactor}{4}
\providecommand{\BIBentryALTinterwordspacing}{\spaceskip=\fontdimen2\font plus
\BIBentryALTinterwordstretchfactor\fontdimen3\font minus
  \fontdimen4\font\relax}
\providecommand{\BIBforeignlanguage}[2]{{%
\expandafter\ifx\csname l@#1\endcsname\relax
\typeout{** WARNING: IEEEtran.bst: No hyphenation pattern has been}%
\typeout{** loaded for the language `#1'. Using the pattern for}%
\typeout{** the default language instead.}%
\else
\language=\csname l@#1\endcsname
\fi
#2}}
\providecommand{\BIBdecl}{\relax}
\BIBdecl

\bibitem{cadenaPresentFutureSimultaneous2016}
C.~Cadena, L.~Carlone, H.~Carrillo, Y.~Latif, D.~Scaramuzza, J.~Neira, I.~Reid,
  and J.~J. Leonard, ``Past, {{Present}}, and {{Future}} of {{Simultaneous
  Localization}} and {{Mapping}}: {{Toward}} the {{Robust-Perception Age}},''
  \emph{IEEE Transactions on Robotics}, vol.~32, no.~6, pp. 1309--1332, Dec.
  2016.

\bibitem{saeediMultipleRobotSimultaneousLocalization2016}
S.~Saeedi, M.~Trentini, M.~Seto, and H.~Li, ``Multiple-{{Robot Simultaneous
  Localization}} and {{Mapping}}: {{A Review}},'' \emph{Journal of Field
  Robotics}, vol.~33, no.~1, pp. 3--46, 2016.

\bibitem{lajoieCollaborativeSimultaneousLocalization2022}
P.-Y. Lajoie, B.~Ramtoula, F.~Wu, and G.~Beltrame, ``Towards {{Collaborative
  Simultaneous Localization}} and {{Mapping}}: A {{Survey}} of the {{Current
  Research Landscape}},'' \emph{Field Robotics}, vol.~2, no.~1, pp. 971--1000,
  Mar. 2022.

\bibitem{anderssonCSAMMultiRobotSLAM2008}
L.~A.~A. Andersson and J.~Nygards, ``C-{{SAM}}: {{Multi-Robot SLAM}} using
  square root information smoothing,'' in \emph{2008 {{IEEE International
  Conference}} on {{Robotics}} and {{Automation}}}, May 2008, pp. 2798--2805.

\bibitem{schmuckCCMSLAMRobustEfficient2019}
P.~Schmuck and M.~Chli, ``{{CCM-SLAM}}: {{Robust}} and efficient centralized
  collaborative monocular simultaneous localization and mapping for robotic
  teams,'' \emph{Journal of Field Robotics}, vol.~36, no.~4, pp. 763--781,
  2019.

\bibitem{lajoieSwarmSLAMSparseDecentralized2024}
P.-Y. Lajoie and G.~Beltrame, ``Swarm-{{SLAM}}: {{Sparse Decentralized
  Collaborative Simultaneous Localization}} and {{Mapping Framework}} for
  {{Multi-Robot Systems}},'' \emph{IEEE Robotics and Automation Letters},
  vol.~9, no.~1, pp. 475--482, Jan. 2024.

\bibitem{CanadianSpaceAgency2021}
``Canadian {{Space Agency Analogue Terrain}},''
  https://www.asc-csa.gc.ca/eng/laboratories-and-warehouse/analogue-terrain.asp,
  Aug. 2021.

\bibitem{lajoieCollaborativeSimultaneousLocalization2024}
P.-Y. Lajoie, ``Collaborative {{Simultaneous Localization}} and {{Mapping
  Dataset}} on {{Mars Analogue Terrain}} with {{Inter-Robot Communication
  Estimates}},'' Dec. 2024.

\bibitem{prorokRobustnessTaxonomyApproaches2021}
A.~Prorok, M.~Malencia, L.~Carlone, G.~S. Sukhatme, B.~M. Sadler, and V.~Kumar,
  ``Beyond {{Robustness}}: {{A Taxonomy}} of {{Approaches}} towards {{Resilient
  Multi-Robot Systems}},'' \emph{arXiv:2109.12343 [cs, eess]}, Sep. 2021.

\bibitem{cieslewskiDataEfficientDecentralizedVisual2018}
T.~Cieslewski, S.~Choudhary, and D.~Scaramuzza, ``Data-{{Efficient
  Decentralized Visual SLAM}},'' in \emph{2018 {{IEEE International
  Conference}} on {{Robotics}} and {{Automation}} ({{ICRA}})}, May 2018, pp.
  2466--2473.

\bibitem{lajoieDOORSLAMDistributedOnline2020}
P.-Y. Lajoie, B.~Ramtoula, Y.~Chang, L.~Carlone, and G.~Beltrame,
  ``{{DOOR-SLAM}}: {{Distributed}}, {{Online}}, and {{Outlier Resilient SLAM}}
  for {{Robotic Teams}},'' \emph{IEEE Robotics and Automation Letters}, vol.~5,
  no.~2, pp. 1656--1663, Apr. 2020.

\bibitem{bertonRethinkingVisualGeoLocalization2022}
G.~Berton, C.~Masone, and B.~Caputo, ``Rethinking {{Visual Geo-Localization}}
  for {{Large-Scale Applications}},'' in \emph{Proceedings of the
  {{IEEE}}/{{CVF Conference}} on {{Computer Vision}} and {{Pattern
  Recognition}}}, 2022, pp. 4878--4888.

\bibitem{kimScanContextEgocentric2018}
G.~Kim and A.~Kim, ``Scan {{Context}}: {{Egocentric Spatial Descriptor}} for
  {{Place Recognition Within 3D Point Cloud Map}},'' in \emph{2018
  {{IEEE}}/{{RSJ International Conference}} on {{Intelligent Robots}} and
  {{Systems}} ({{IROS}})}, Oct. 2018, pp. 4802--4809.

\bibitem{yangTEASERFastCertifiable2021}
H.~Yang, J.~Shi, and L.~Carlone, ``{{TEASER}}: {{Fast}} and {{Certifiable Point
  Cloud Registration}},'' \emph{IEEE Transactions on Robotics}, vol.~37, no.~2,
  pp. 314--333, Apr. 2021.

\bibitem{choudharyDistributedMappingPrivacy2017}
S.~Choudhary, L.~Carlone, C.~Nieto, J.~Rogers, H.~I. Christensen, and
  F.~Dellaert, ``Distributed mapping with privacy and communication
  constraints: {{Lightweight}} algorithms and object-based models,'' \emph{The
  International Journal of Robotics Research}, vol.~36, no.~12, pp. 1286--1311,
  Oct. 2017.

\bibitem{tianDistributedCertifiablyCorrect2021}
Y.~Tian, K.~Khosoussi, D.~M. Rosen, and J.~P. How, ``Distributed {{Certifiably
  Correct Pose-Graph Optimization}},'' \emph{IEEE Transactions on Robotics},
  pp. 1--20, 2021.

\bibitem{muraiRobotWebDistributed2024}
R.~Murai, J.~Ortiz, S.~Saeedi, P.~H.~J. Kelly, and A.~J. Davison, ``A {{Robot
  Web}} for {{Distributed Many-Device Localization}},'' \emph{IEEE Transactions
  on Robotics}, vol.~40, pp. 121--138, 2024.

\bibitem{fanMajorizationMinimizationMethods2024}
T.~Fan and T.~D. Murphey, ``Majorization {{Minimization Methods}} for
  {{Distributed Pose Graph Optimization}},'' \emph{IEEE Transactions on
  Robotics}, vol.~40, pp. 22--42, 2024.

\bibitem{mcgannIMESAIncrementalDistributed2024b}
D.~McGann and M.~Kaess, ``{{iMESA}}: {{Incremental Distributed Optimization}}
  for {{Collaborative Simultaneous Localization}} and {{Mapping}},'' in
  \emph{Robotics: {{Science}} and {{Systems XX}}}.\hskip 1em plus 0.5em minus
  0.4em\relax {Robotics: Science and Systems Foundation}, Jul. 2024.

\bibitem{lajoieModelingPerceptualAliasing2019}
P.-Y. Lajoie, S.~Hu, G.~Beltrame, and L.~Carlone, ``Modeling {{Perceptual
  Aliasing}} in {{SLAM}} via {{Discrete}}--{{Continuous Graphical Models}},''
  \emph{IEEE Robotics and Automation Letters}, vol.~4, no.~2, pp. 1232--1239,
  Apr. 2019.

\bibitem{ebadiLAMPLargeScaleAutonomous2020}
K.~Ebadi, Y.~Chang, M.~Palieri, A.~Stephens, A.~Hatteland, E.~Heiden,
  A.~Thakur, N.~Funabiki, B.~Morrell, S.~Wood, L.~Carlone, and A.-a.
  {Agha-mohammadi}, ``{{LAMP}}: {{Large-Scale Autonomous Mapping}} and
  {{Positioning}} for {{Exploration}} of {{Perceptually-Degraded Subterranean
  Environments}},'' in \emph{2020 {{IEEE International Conference}} on
  {{Robotics}} and {{Automation}} ({{ICRA}})}, May 2020, pp. 80--86.

\bibitem{tianSearchRescueForest2020}
Y.~Tian, K.~Liu, K.~Ok, L.~Tran, D.~Allen, N.~Roy, and J.~P. How, ``Search and
  rescue under the forest canopy using multiple {{UAVs}},'' \emph{The
  International Journal of Robotics Research}, vol.~39, no. 10-11, pp.
  1201--1221, Sep. 2020.

\bibitem{mangelsonPairwiseConsistentMeasurement2018}
J.~G. Mangelson, D.~Dominic, R.~M. Eustice, and R.~Vasudevan, ``Pairwise
  {{Consistent Measurement Set Maximization}} for {{Robust Multi-Robot Map
  Merging}},'' in \emph{2018 {{IEEE International Conference}} on {{Robotics}}
  and {{Automation}} ({{ICRA}})}, May 2018, pp. 2916--2923.

\bibitem{yangGraduatedNonConvexityRobust2020}
H.~Yang, P.~Antonante, V.~Tzoumas, and L.~Carlone, ``Graduated
  {{Non-Convexity}} for {{Robust Spatial Perception}}: {{From Non-Minimal
  Solvers}} to {{Global Outlier Rejection}},'' \emph{IEEE Robotics and
  Automation Letters}, vol.~5, no.~2, pp. 1127--1134, Apr. 2020.

\bibitem{tianKimeraMultiRobustDistributed2022}
Y.~Tian, Y.~Chang, F.~Herrera~Arias, C.~{Nieto-Granda}, J.~P. How, and
  L.~Carlone, ``Kimera-{{Multi}}: {{Robust}}, {{Distributed}}, {{Dense
  Metric-Semantic SLAM}} for {{Multi-Robot Systems}},'' \emph{IEEE Transactions
  on Robotics}, vol.~38, no.~4, pp. 2022--2038, Aug. 2022.

\bibitem{sarlinSuperGlueLearningFeature2020}
P.-E. Sarlin, D.~DeTone, T.~Malisiewicz, and A.~Rabinovich, ``{{SuperGlue}}:
  {{Learning Feature Matching With Graph Neural Networks}},'' in \emph{2020
  {{IEEE}}/{{CVF Conference}} on {{Computer Vision}} and {{Pattern
  Recognition}} ({{CVPR}})}, Jun. 2020, pp. 4937--4946.

\bibitem{ebadiRoverLocalizationMars2020}
K.~Ebadi and A.-A. {Agha-Mohammadi}, ``Rover {{Localization}} in {{Mars
  Helicopter Aerial Maps}}: {{Experimental Results}} in a {{Mars-Analogue
  Environment}},'' in \emph{Proceedings of the 2018 {{International Symposium}}
  on {{Experimental Robotics}}}, J.~Xiao, T.~Kr{\"o}ger, and O.~Khatib,
  Eds.\hskip 1em plus 0.5em minus 0.4em\relax Cham: Springer International
  Publishing, 2020, pp. 72--84.

\bibitem{newmanNavigatingRecognizingDescribing2009}
P.~Newman, G.~Sibley, M.~Smith, M.~Cummins, A.~Harrison, C.~Mei, I.~Posner,
  R.~Shade, D.~Schroeter, L.~Murphy, W.~Churchill, D.~Cole, and I.~Reid,
  ``Navigating, {{Recognizing}} and {{Describing Urban Spaces With Vision}} and
  {{Lasers}},'' \emph{The International Journal of Robotics Research}, vol.~28,
  no. 11-12, pp. 1406--1433, Nov. 2009.

\bibitem{shanLVISAMTightlycoupledLidarVisualInertial2021}
T.~Shan, B.~Englot, C.~Ratti, and D.~Rus, ``{{LVI-SAM}}: {{Tightly-coupled
  Lidar-Visual-Inertial Odometry}} via {{Smoothing}} and {{Mapping}},'' in
  \emph{2021 {{IEEE International Conference}} on {{Robotics}} and
  {{Automation}} ({{ICRA}})}, May 2021, pp. 5692--5698.

\bibitem{huangDiSCoSLAMDistributedScan2022}
Y.~Huang, T.~Shan, F.~Chen, and B.~Englot, ``{{DiSCo-SLAM}}: {{Distributed Scan
  Context-Enabled Multi-Robot LiDAR SLAM With Two-Stage Global-Local Graph
  Optimization}},'' \emph{IEEE Robotics and Automation Letters}, vol.~7, no.~2,
  pp. 1150--1157, Apr. 2022.

\bibitem{liuSlideSLAMSparseLightweight2024}
X.~Liu, J.~Lei, A.~Prabhu, Y.~Tao, I.~Spasojevic, P.~Chaudhari, N.~Atanasov,
  and V.~Kumar, ``{{SlideSLAM}}: {{Sparse}}, {{Lightweight}}, {{Decentralized
  Metric-Semantic SLAM}} for {{Multi-Robot Navigation}},'' Dec. 2024.

\bibitem{schmuckCOVINSVisualInertialSLAM2021}
P.~Schmuck, T.~Ziegler, M.~Karrer, J.~Perraudin, and M.~Chli, ``{{COVINS}}:
  {{Visual-Inertial SLAM}} for {{Centralized Collaboration}},'' in \emph{2021
  {{IEEE International Symposium}} on {{Mixed}} and {{Augmented Reality
  Adjunct}} ({{ISMAR-Adjunct}})}, Oct. 2021, pp. 171--176.

\bibitem{changLAMP20Robust2022}
Y.~Chang, K.~Ebadi, C.~E. Denniston, M.~F. Ginting, A.~Rosinol, A.~Reinke,
  M.~Palieri, J.~Shi, A.~Chatterjee, B.~Morrell, A.-a. {Agha-mohammadi}, and
  L.~Carlone, ``{{LAMP}} 2.0: {{A Robust Multi-Robot SLAM System}} for
  {{Operation}} in {{Challenging Large-Scale Underground Environments}},''
  \emph{IEEE Robotics and Automation Letters}, vol.~7, no.~4, pp. 9175--9182,
  Oct. 2022.

\bibitem{cramariucMaplab20Modular2023}
A.~Cramariuc, L.~Bernreiter, F.~Tschopp, M.~Fehr, V.~Reijgwart, J.~Nieto,
  R.~Siegwart, and C.~Cadena, ``Maplab 2.0 -- {{A Modular}} and {{Multi-Modal
  Mapping Framework}},'' \emph{IEEE Robotics and Automation Letters}, vol.~8,
  no.~2, pp. 520--527, Feb. 2023.

\bibitem{schusterARCHESSpaceAnalogueDemonstration2020}
M.~J. Schuster, M.~G. Muller, S.~G. Brunner, H.~Lehner, P.~Lehner, R.~Sakagami,
  A.~Domel, L.~Meyer, B.~Vodermayer, R.~Giubilato, M.~Vayugundla, J.~Reill,
  F.~Steidle, I.~Von~Bargen, K.~Bussmann, R.~Belder, P.~Lutz, W.~Sturzl,
  M.~Smisek, M.~Maier, S.~Stoneman, A.~F. Prince, B.~Rebele, M.~Durner,
  E.~Staudinger, S.~Zhang, R.~Pohlmann, E.~Bischoff, C.~Braun, S.~Schroder,
  E.~Dietz, S.~Frohmann, A.~Borner, H.-W. Hubers, B.~Foing, R.~Triebel, A.~O.
  {Albu-Schaffer}, and A.~Wedler, ``The {{ARCHES Space-Analogue Demonstration
  Mission}}: {{Towards Heterogeneous Teams}} of {{Autonomous Robots}} for
  {{Collaborative Scientific Sampling}} in {{Planetary Exploration}},''
  \emph{IEEE Robotics and Automation Letters}, vol.~5, no.~4, pp. 5315--5322,
  Oct. 2020.

\bibitem{schusterDistributedStereoVisionbased2019a}
M.~J. Schuster, K.~Schmid, C.~Brand, and M.~Beetz, ``Distributed stereo
  vision-based {{6D}} localization and mapping for multi-robot teams,''
  \emph{Journal of Field Robotics}, vol.~36, no.~2, pp. 305--332, 2019.

\bibitem{shengMinimizingDataExchange2005}
W.~Sheng, Q.~Wang, Q.~Yang, and S.~Zhu, ``Minimizing data exchange in ad hoc
  multi-robot networks,'' in \emph{{{ICAR}} '05. {{Proceedings}}., 12th
  {{International Conference}} on {{Advanced Robotics}}, 2005.}, Jul. 2005, pp.
  811--816.

\bibitem{varadharajanSOULDataSharing2020}
V.~S. Varadharajan, D.~{St-Onge}, B.~Adams, and G.~Beltrame, ``{{SOUL}}: Data
  sharing for robot swarms,'' \emph{Autonomous Robots}, vol.~44, no.~3, pp.
  377--394, Mar. 2020.

\bibitem{ghediniEfficientAdaptiveAdhoc2018}
C.~Ghedini, C.~H.~C. Ribeiro, and L.~Sabattini, ``Toward efficient adaptive
  ad-hoc multi-robot network topologies,'' \emph{Ad Hoc Networks}, vol.~74, pp.
  57--70, May 2018.

\bibitem{siligardiRobustAreaCoverage2019}
L.~Siligardi, J.~Panerati, M.~Kaufmann, M.~Minelli, C.~Ghedini, G.~Beltrame,
  and L.~Sabattini, ``Robust {{Area Coverage}} with {{Connectivity
  Maintenance}},'' in \emph{2019 {{International Conference}} on {{Robotics}}
  and {{Automation}} ({{ICRA}})}, May 2019, pp. 2202--2208.

\bibitem{varadharajanSwarmRelaysDistributed2020}
V.~S. Varadharajan, D.~{St-Onge}, B.~Adams, and G.~Beltrame, ``Swarm
  {{Relays}}: {{Distributed Self-Healing Ground-and-Air Connectivity
  Chains}},'' \emph{IEEE Robotics and Automation Letters}, vol.~5, no.~4, pp.
  5347--5354, Oct. 2020.

\bibitem{giamouTalkResourceEfficientlyMe2018}
M.~Giamou, K.~Khosoussi, and J.~P. How, ``Talk {{Resource-Efficiently}} to
  {{Me}}: {{Optimal Communication Planning}} for {{Distributed Loop Closure
  Detection}},'' in \emph{2018 {{IEEE International Conference}} on
  {{Robotics}} and {{Automation}} ({{ICRA}})}, May 2018, pp. 3841--3848.

\bibitem{tianOptimalBudgetedData2018}
Y.~Tian, K.~Khosoussi, M.~Giamou, J.~How, and J.~Kelly, ``Near-{{Optimal
  Budgeted Data Exchange}} for {{Distributed Loop Closure Detection}},'' in
  \emph{Robotics: {{Science}} and {{Systems XIV}}}.\hskip 1em plus 0.5em minus
  0.4em\relax {Robotics: Science and Systems Foundation}, Jun. 2018.

\bibitem{tianResilientDistributedMultiRobot2023}
Y.~Tian, Y.~Chang, L.~Quang, A.~Schang, C.~{Nieto-Granda}, J.~P. How, and
  L.~Carlone, ``Resilient and {{Distributed Multi-Robot Visual SLAM}}:
  {{Datasets}}, {{Experiments}}, and {{Lessons Learned}},'' in \emph{2023
  {{IEEE}}/{{RSJ International Conference}} on {{Intelligent Robots}} and
  {{Systems}} ({{IROS}})}, Oct. 2023, pp. 11\,027--11\,034.

\bibitem{leungUTIASMultirobotCooperative2011}
K.~Y. Leung, Y.~Halpern, T.~D. Barfoot, and H.~H. Liu, ``The {{UTIAS}}
  multi-robot cooperative localization and mapping dataset,'' \emph{The
  International Journal of Robotics Research}, vol.~30, no.~8, pp. 969--974,
  Jul. 2011.

\bibitem{duboisAirMuseumHeterogeneousMultirobot2020}
R.~Dubois, A.~Eudes, and V.~Fr{\'e}mont, ``{{AirMuseum}}: A heterogeneous
  multi-robot dataset for stereo-visual and inertial {{Simultaneous
  Localization And Mapping}},'' in \emph{2020 {{IEEE International Conference}}
  on {{Multisensor Fusion}} and {{Integration}} for {{Intelligent Systems}}
  ({{MFI}})}, Sep. 2020, pp. 166--172.

\bibitem{zhuGRACOMultimodalDataset2023}
Y.~Zhu, Y.~Kong, Y.~Jie, S.~Xu, and H.~Cheng, ``{{GRACO}}: {{A Multimodal
  Dataset}} for {{Ground}} and {{Aerial Cooperative Localization}} and
  {{Mapping}},'' \emph{IEEE Robotics and Automation Letters}, vol.~8, no.~2,
  pp. 966--973, Feb. 2023.

\bibitem{fengS3EMultiRobotMultimodal2024}
D.~Feng, ``{{S3E}}: {{A Multi-Robot Multimodal Dataset}} for {{Collaborative
  SLAM}},'' Jul. 2024.

\bibitem{parkBenchmarkDatasetCollaborative2024}
H.~Park, I.~Lee, M.~Kim, H.~Park, and K.~Joo, ``A {{Benchmark Dataset}} for
  {{Collaborative SLAM}} in {{Service Environments}},'' \emph{IEEE Robotics and
  Automation Letters}, vol.~9, no.~12, pp. 11\,337--11\,344, Dec. 2024.

\bibitem{zhaoSubTMRSDatasetPushing2024}
S.~Zhao, Y.~Gao, T.~Wu, D.~Singh, R.~Jiang, H.~Sun, M.~Sarawata, Y.~Qiu,
  W.~Whittaker, I.~Higgins, Y.~Du, S.~Su, C.~Xu, J.~Keller, J.~Karhade,
  L.~Nogueira, S.~Saha, J.~Zhang, W.~Wang, C.~Wang, and S.~Scherer,
  ``{{SubT-MRS Dataset}}: {{Pushing SLAM Towards All-weather Environments}},''
  May 2024.

\bibitem{shanLIOSAMTightlycoupledLidar2020}
T.~Shan, B.~Englot, D.~Meyers, W.~Wang, C.~Ratti, and D.~Rus, ``{{LIO-SAM}}:
  {{Tightly-coupled Lidar Inertial Odometry}} via {{Smoothing}} and
  {{Mapping}},'' in \emph{2020 {{IEEE}}/{{RSJ International Conference}} on
  {{Intelligent Robots}} and {{Systems}} ({{IROS}})}, Oct. 2020, pp.
  5135--5142.

\bibitem{fainelli2008openwrt}
F.~Fainelli, ``The openwrt embedded development framework,'' in
  \emph{Proceedings of the Free and Open Source Software Developers European
  Meeting}, 2008, p. 106.

\bibitem{corsaro2023zenoh}
A.~Corsaro, L.~Cominardi, O.~Hecart, G.~Baldoni, J.~E.~P. Avital, J.~Loudet,
  C.~Guimares, M.~Ilyin, and D.~Bannov, ``Zenoh: Unifying communication,
  storage and computation from the cloud to the microcontroller,'' in
  \emph{2023 26th Euromicro Conference on Digital System Design (DSD)}.\hskip
  1em plus 0.5em minus 0.4em\relax IEEE, 2023, pp. 422--428.

\bibitem{dunganIPerfTCPUDP}
J.~Dungan, S.~Elliot, B.~A. Mah, J.~Poskanzer, and P.~Kaustubh, ``{{iPerf}} -
  {{The TCP}}, {{UDP}} and {{SCTP}} network bandwidth measurement tool,''
  https://iperf.fr/.

\bibitem{schemersFPing}
R.~Schemers, ``{{fPing}},'' https://fping.org/.

\bibitem{gruppEvoPythonPackage2017}
M.~Grupp, ``Evo: {{Python}} package for the evaluation of odometry and
  {{SLAM}}.'' 2017.

\bibitem{schlagerAutomotiveLidarVibration2022}
B.~Schlager, T.~Goelles, M.~Behmer, S.~Muckenhuber, J.~Payer, and D.~Watzenig,
  ``Automotive {{Lidar}} and {{Vibration}}: {{Resonance}}, {{Inertial
  Measurement Unit}}, and {{Effects}} on the {{Point Cloud}},'' \emph{IEEE Open
  Journal of Intelligent Transportation Systems}, vol.~3, pp. 426--434, 2022.

\bibitem{kegeleirsSwarmSLAMChallenges2021}
M.~Kegeleirs, G.~Grisetti, and M.~Birattari, ``Swarm {{SLAM}}: {{Challenges}}
  and {{Perspectives}},'' \emph{Frontiers in Robotics and AI}, vol.~8, 2021.

\bibitem{houshiar3DPointCloud2015}
H.~Houshiar and A.~N{\"u}chter, ``{{3D}} point cloud compression using
  conventional image compression for efficient data transmission,'' in
  \emph{2015 {{XXV International Conference}} on {{Information}},
  {{Communication}} and {{Automation Technologies}} ({{ICAT}})}, Oct. 2015, pp.
  1--8.

\bibitem{han3DPointCloud2023}
X.-F. Han, Z.-A. Feng, S.-J. Sun, and G.-Q. Xiao, ``{{3D}} point cloud
  descriptors: State-of-the-art,'' \emph{Artificial Intelligence Review},
  vol.~56, no.~10, pp. 12\,033--12\,083, Oct. 2023.

\bibitem{xuVoxelbasedRepresentation3D2021}
Y.~Xu, X.~Tong, and U.~Stilla, ``Voxel-based representation of {{3D}} point
  clouds: {{Methods}}, applications, and its potential use in the construction
  industry,'' \emph{Automation in Construction}, vol. 126, p. 103675, Jun.
  2021.

\bibitem{changHydraMultiCollaborativeOnline2023}
Y.~Chang, N.~Hughes, A.~Ray, and L.~Carlone, ``Hydra-{{Multi}}: {{Collaborative
  Online Construction}} of {{3D Scene Graphs}} with {{Multi-Robot Teams}},'' in
  \emph{2023 {{IEEE}}/{{RSJ International Conference}} on {{Intelligent
  Robots}} and {{Systems}} ({{IROS}})}, Oct. 2023, pp. 10\,995--11\,002.

\end{thebibliography}

\vfill\pagebreak

\end{document}